\documentclass{article}

\usepackage[final]{corl_2019} % Uncomment for the camera-ready ``final'' version

\usepackage[numbers]{natbib}
\usepackage{multicol}
\usepackage{subfigure}
\usepackage{graphicx}
\usepackage{comment}
\usepackage{balance}
\usepackage{amsmath}
\usepackage{color}

\title{Understanding Teacher Gaze Patterns\\ for Robot Learning}

\author{
  Akanksha Saran\\
  Department of Computer Science\\
  University of Texas at Austin\\
  \texttt{asaran@cs.utexas.edu} \\
  \And
  Elaine Schaertl Short\\
  Electrical and Computer Engineering\\
  University of Texas at Austin\\
  \texttt{elaine.short@utexas.edu} \\
  \AND
  Andrea Thomaz\\
  Electrical and Computer Engineering\\
  University of Texas at Austin\\
  \texttt{athomaz@ece.utexas.edu} \\
  \And
  Scott Niekum\\
  Department of Computer Science\\
  University of Texas at Austin\\
  \texttt{sniekum@cs.utexas.edu} \\
}

\begin{document}
\maketitle

%===============================================================================

\begin{abstract}
 Human gaze is known to be a strong indicator of underlying human intentions and goals during manipulation tasks. This work studies gaze patterns of human teachers demonstrating tasks to robots and proposes ways in which such patterns can be used to enhance robot learning. 
Using both kinesthetic teaching and video demonstrations, we identify novel intention-revealing gaze behaviors during teaching. These prove to be informative in a variety of problems ranging from reference frame inference to segmentation of multi-step tasks.
Based on our findings, we propose two proof-of-concept algorithms which show that gaze data can enhance subtask classification for a multi-step task up to 6\% and reward inference and policy learning for a single-step task up to 67\%. 
Our findings provide a foundation for a model of natural human gaze in robot learning from demonstration settings and present open problems for utilizing human gaze to enhance robot learning. 
\end{abstract}

% Two or three meaningful keywords should be added here
\keywords{Learning from demonstrations, Eye gaze, Kinesthetic Teaching, Learning from observations} 

%===============================================================================

\section{Introduction}
Eye gaze is an important social cue that humans use  to convey goals, future actions, and mental load \cite{argyle1976gaze, argyle1972non} both in verbal and non-verbal settings. In a teacher-learner setup, parents can scaffold a child's learning process by directing their attention using gaze, thereby providing structure to the task \cite{trafton2008integrating, wertsch1984creation}. 
As in human-human interactions, we hypothesize that gaze can play a role in guiding robot learning from humans. To understand what role human gaze plays when humans teach robots, we study eye gaze behaviors in the context of robot learning from demonstrations (LfD) \cite{argall2009survey}, a powerful, natural framework that allows non-experts to communicate rich task knowledge to robots by showing them how to perform a task. We focus on two modalities of LfD for robot manipulation \cite{kroemer2019review}: (1) learning via keyframe-based kinesthetic teaching (KT) in which the joints of a robot are moved by a human teacher through specific points or keyframes while the robot records its joint configurations at these keyframes \cite{akgun2012trajectories}, and (2) learning from observation, specifically video demonstrations, in which a robot can passively observe a human performing the task and learn how the demonstrated actions translate to its own body to achieve the same goal. Video demonstrations are often freely available on the web for many skills needed in offices or households by robots, which makes them a popular choice for robot learning. Learning algorithms for these techniques typically use trajectories of state-action pairs directly or indirectly. In addition to knowledge about actions, information about teacher intent in the form of eye gaze can enhance learning from demonstrations in terms of generalizing to new environments and learning with fewer demonstrations. 

To use eye gaze for LfD algorithms, it is necessary to understand gaze behavior during the interaction for a specific demonstration type. The psychological literature has characterized gaze behavior of people performing certain manipulation tasks with their own hands, like moving objects around obstacles \cite{johansson2001eye} or making tea \cite{land1999eye}. These studies show that gaze follows the objects involved in the task \cite{hayhoe2005eye} and that eye gaze precedes hand motion \cite{johansson2001eye, land2001ways}. These insights can be applicable to video demonstrations for a robot. However, our work is the first to study eye gaze for paired kinesthetic teaching (KT) and video demonstrations with an eye toward ways it can be used computationally. Thus in this work, we aim to characterize the gaze behavior of human teachers demonstrating the same task under both teaching paradigms to a robot. We perform a data collection study in which human subjects wear an eye tracker to provide accurate ground truth for gaze fixations. In practice, our findings should be useful even without access to a gaze tracker, through the use of vision based algorithms to predict gaze fixations \citep{recasens2015they, chong2018connecting}.

In our study, we find that users spend most of their time fixating on objects which are relevant for completing the task. Moreover, human gaze can reveal information about the human's intentions in otherwise ambiguous situations and help predict the reference frame with respect to which certain keyframes or actions are demonstrated.  We also show that human gaze is a meaningful feature to distinguish between keyframes which demarcate semantically different actions (step keyframes) versus contiguous keyframes which belong to the same semantic action (non-step keyframes) such as multiple keyframes shaping a pouring motion. These insights open up exciting new avenues for research in learning from human gaze such as automatic segmentation of a task into subtasks, inferring intentions and goals of such subtasks, and efficient robot learning. We observe up to 6\% improvement in subtask classification with gaze during a multi-step task for both demonstration types. As another potential application, we show improved goal inference and policy learning for a manipulation task by augmenting Bayesian inverse reinforcement learning with gaze information from the demonstrator. Policy loss improves by 67.4\% for video demonstrations compared to 53.75\% for KT demonstrations, suggesting that video demonstrations are a richer and more compact source of intention-revealing gaze signals.

%===============================================================================
\section{Related Work}

Human gaze and attention are known to be task-dependent and goal-oriented \cite{yarbus1967eye}. 
\citet{flanagan2003action} demonstrated that adults predict action goals by fixating on the end location of an action before it is reached, both when they execute an action themselves and when they observe someone else executing it. Single fixations have identifiable functions (locating, directing, guiding, and checking) related to the action to be taken. \citet{hayhoe2005eye} show that the point of fixation in a given scenario may not be the most visually salient location, but rather corresponds to a location important for the specifications and spatio-temporal demands of the task. This line of investigation has been used in extended visuo-motor tasks such as driving, walking, sports, and making tea or sandwiches \cite{hayhoe2000vision, hayhoe2003visual}. It has also been found that eye gaze fixations are tightly linked in time to the evolution of the task and very few irrelevant areas are fixated upon \cite{land2009vision}, implying that the control of fixation comes principally from top-down instructions, not bottom-up salience. Tasks such as block copying \cite{ballard1992hand} and block manipulation \cite{johansson2001eye} have shown that gaze was directed to locations where information critical for manipulation was obtained. 
Subjects appear to use gaze to select specific information required at a specific point of time in a block manipulation task \cite{johansson2001eye, ballard1995memory}. \citet{ballard1995memory} referred to this efficient effect as a `just-in-time' strategy, as only a limited amount of information needs to be computed from the visual image within a fixation, and it is not necessary to maintain this information in working memory if it is no longer needed. 
These studies suggest that gaze would be helpful in predicting the intention or goal location of human manipulation actions. In our work, we study such human gaze patterns for paired video and KT demonstrations and recover characteristics specific to demonstrations for robots.

There  is  also a  rich  body  of  work  on  eye  gaze  for  human-robot  interaction \cite{admoni2017social}.  \citet{hart2014gesture} use nonverbal cues including gaze to study timing coordination between humans and robots.  Gaze  information  has also been shown to enable  the  establishment  of joint  attention  between  the  human  and  robot  partner,  the recognition  of  human  behavior  and  the  execution  of anticipatory  actions \cite{admoni2017social}.  However, these prior works focus on gaze cues generated by the robot and not on gaze cues from humans. More recently, \citet{aronson2018eye} studied human gaze behavior for shared manipulation, where users controlled a robot arm mounted on a wheelchair via a joystick for assistive tasks of daily living. Novel patterns of gaze behaviors were identified, such as people using visual feedback for aligning the robot arm in a certain orientation and cognitive load being higher for teleoperation versus the shared autonomy condition. However, eye gaze behavior of human teachers has not been studied in the context of robot learning from demonstrations.

Prior research in computer vision has established that task and activity recognition in egocentric videos (similar to video demonstrations in our setup) can benefit from human gaze \citep{tavakoli2019digging, lu2019deep, fathi2012learning, huang2019mutual, shapovalova2013action}. While some of these works predict human gaze as an intermediate output of a deep network classifying human activities \citep{lu2019deep, huang2019mutual}, others either jointly predict gaze and action labels with a probabilistic generative model explicitly modeling properties of gaze behavior \citep{fathi2012learning} or use human gaze as a weak supervisory signal in a latent SVM learning framework \citep{shapovalova2013action}. By contrast, we show a proof of concept for gaze aiding classification of subtasks or actions in egocentric videos of \textit{both} video and KT demonstrations for a multi-step task, which has the potential to further enable task segmentation and policy learning per step. 

There has  also been some recent work on utilizing human eye gaze for learning algorithms. \citet{penkov2017physical}  used  demonstrations  from  a person wearing an eye tracking hardware along with an  egocentric  camera  to  simultaneously ground  symbols  to their instances in the environment and learn the appearance of such object instances. \citet{ravichandar2018gaze} use gaze information as a heuristic to compute a prior distribution of the goal location for reaching motions in a manipulation task. This allows for efficient inference of a multiple-model filtering approach for early intention recognition of reaching actions by pruning model-matching filters that need to be run in parallel. In our work, we show that the use of gaze in conjunction with state-action knowledge can improve reward learning via Bayesian inverse reinforcement learning (BIRL) \cite{ramachandran2007bayesian}.

%===============================================================================
\section{Data Collection And Analysis}
\subsection{User Study Design}
\begin{figure}
\centering
%\vspace{2mm}
\subfigure[Pouring Task]{
\includegraphics[width=0.23\textwidth]{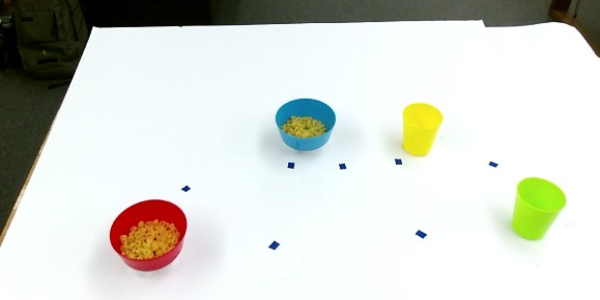}
}
\subfigure[Placement Task]{
\includegraphics[width=0.23\textwidth]{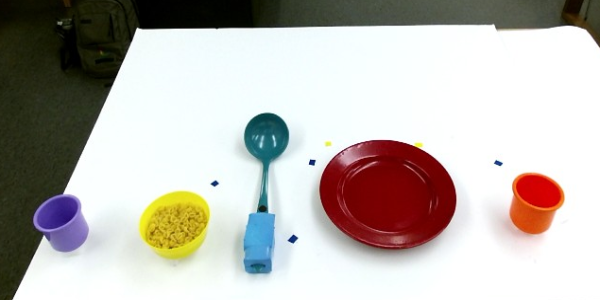}
}
\subfigure[Third person view of 2 demonstration modalities]{
\includegraphics[width=0.23\linewidth]{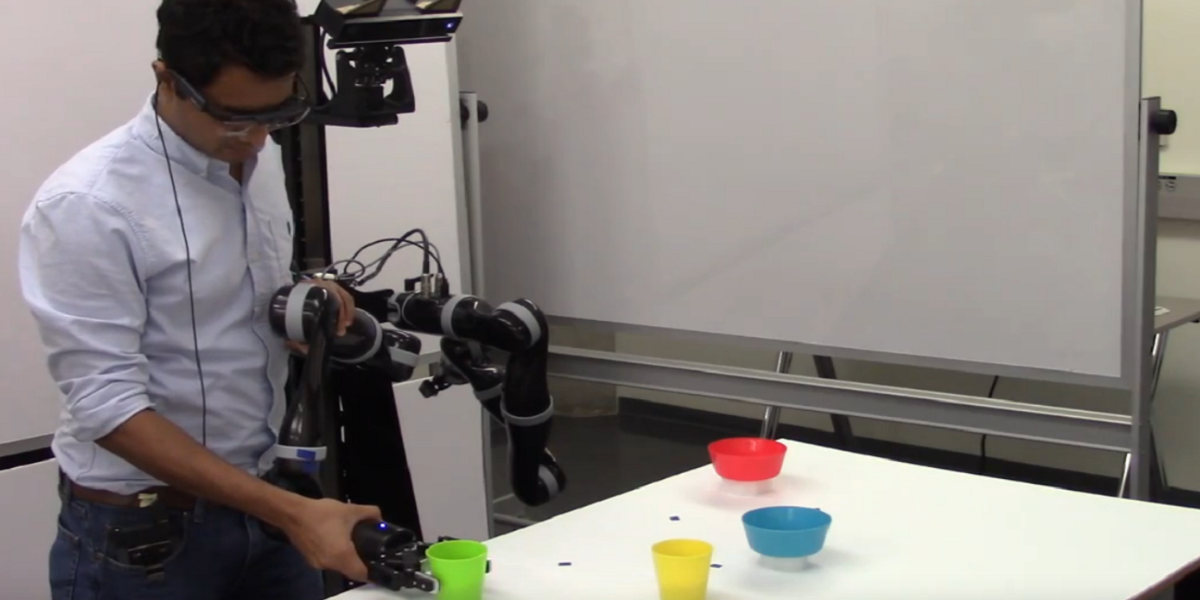}
\includegraphics[width=0.23\linewidth]{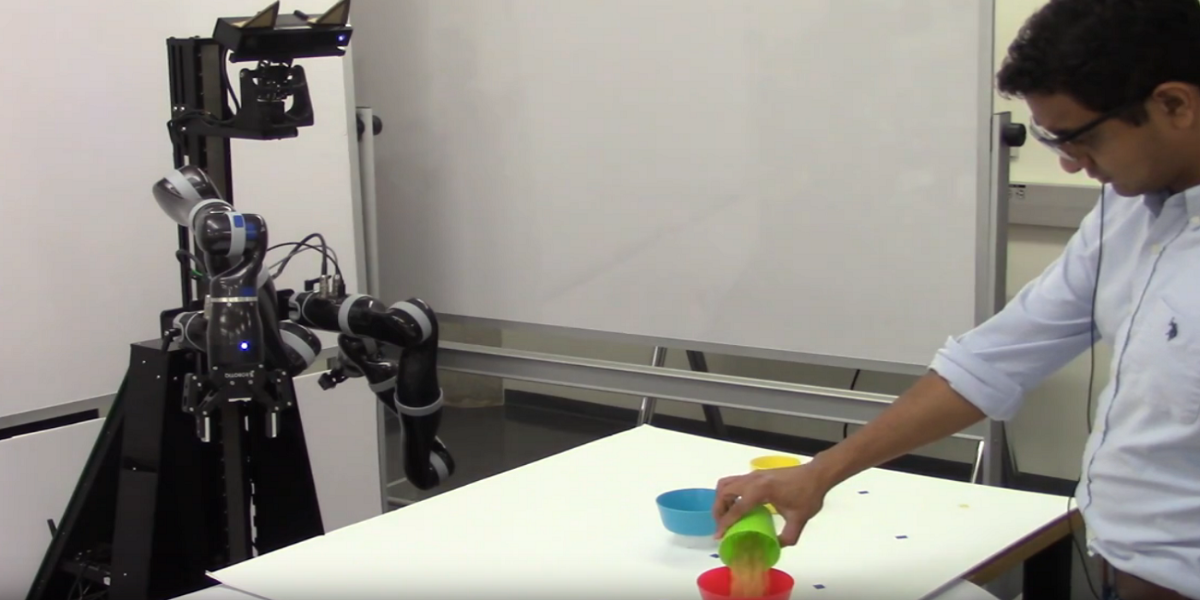}
}
\caption{Task completion configurations for: (a) Pouring Task - where pasta from the green cup is poured into the red bowl and from the yellow cup into the blue bowl; (b) Placement Task - where the green ladle is placed either to the right of the yellow bowl or to the left of the red plate (note both instructions refer to the same ambiguous location). (c) A third person view of a KT and a video demonstration provided by the same user.}
\label{fig:tasks}
\end{figure}

We designed a two-way $2 \times 2$ mixed-design human subjects study (user type: novice or expert $\times$ gaze fixation area: task relevant objects or task-irrelevant object/area) for two household tasks relevant to personal robots: pouring and placement. The task layouts, kept the same across all users, are shown in Fig.\ref{fig:tasks} (a), (b). The details of the two goal-directed tasks are as follows:\\
    \textbf{Pouring task:} Two cups (green and yellow) filled with pasta and two empty bowls (blue and red) are placed at pre-specified locations on a table in front of the robot. Users pour pasta from the green cup to the red bowl, followed by pouring from the yellow cup to the blue bowl. We ask users to provide 3 demonstrations of each type for this task, where demonstration type is counter-balanced across users.\\
    \textbf{Placement task:} Four objects are placed on the table in pre-determined locations -- a purple cup, a yellow bowl filled with yellow pasta, a red plate and an orange cup. A large green ladle with a light blue foam support on the edge of its handle (to ease grasping by the robot gripper), is held in the hand of the demonstrator for a video demonstration or gripped by the robot for the kinesthetic demonstration. The user is given an instruction to place the spoon on a relative location on the table with respect to other objects. Each user is given two instructions (order of instructions is counter-balanced across users) for each demonstration type -- (1) place the green ladle to the left side of the red plate, and (2) place the green ladle to the right side of the yellow bowl. The red plate and the yellow bowl are adjacent to one another on the table with a gap in between them to place the spoon. If the ladle is placed between these two objects on the table, it can be ambiguous to determine which instruction was followed by the user. There were 4 demonstrations provided by each user in this task (2 instructions x 2 demonstration types).

The order of demonstration types were counterbalanced across users. We recruited 20 participants (14 males, 6 females): 10 expert users who had operated or worked with a robot arm, and 10 novice users who had no prior experience operating a robot.  
Each participant was allowed one practice round for each demonstration type in the order they were assigned the demonstration type. After one round of practicing, participants completed 6 demonstrations (3 KT, 3 video) for the pouring task first and then 4 demonstrations (2 KT, 2 video) for the placement task. A trial of a video demonstration lasted between 1.5 seconds to about 25 seconds, and a KT demonstration lasted between 2 minutes to 7 minutes.

%===============================================================================

\begin{figure*}
\centering
\subfigure[Reach %\newline Target:Green Cup
]{
\includegraphics[width=0.147\linewidth]{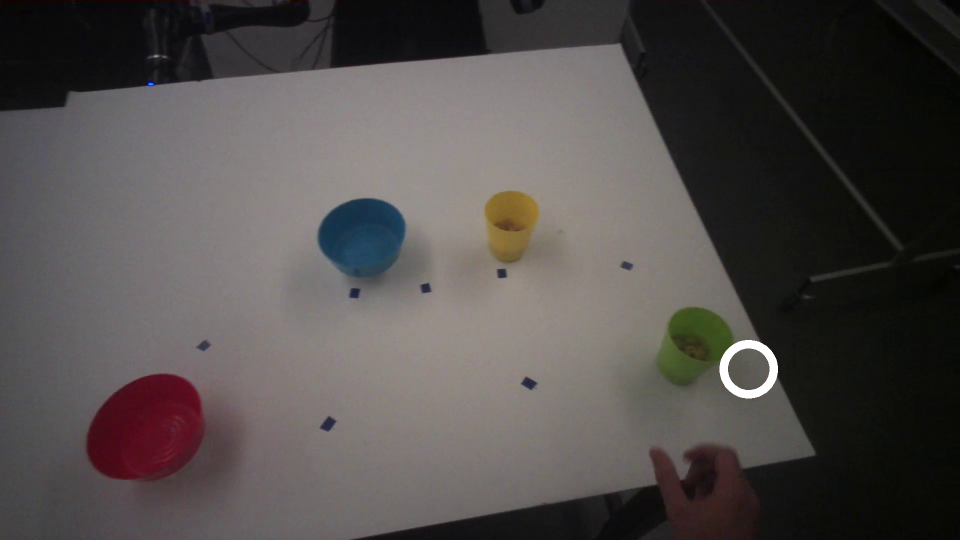}
}
\subfigure[Grasp %\newline Target:Green Cup
]{
\includegraphics[width=0.147\linewidth]{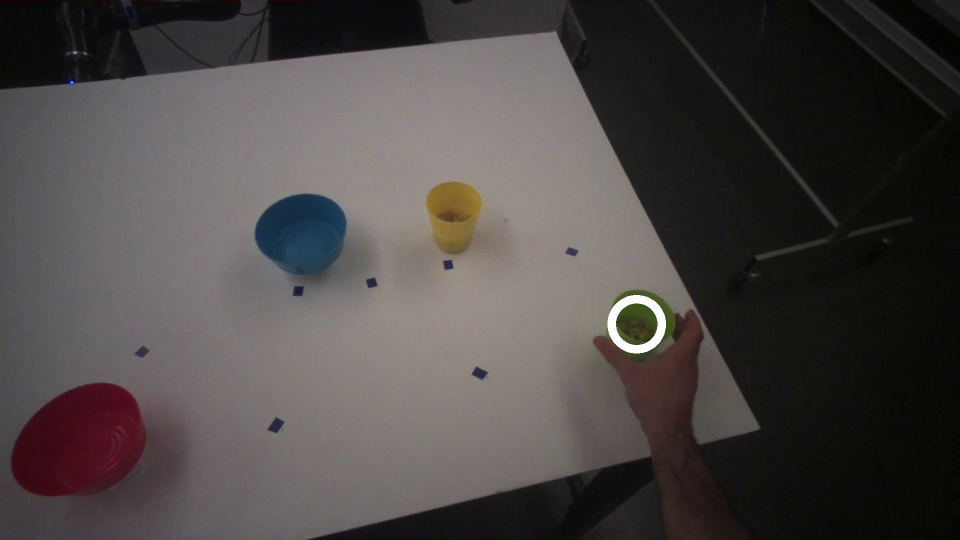}
}
\subfigure[Transport %\newline Target:Red Bowl
]{
\includegraphics[width=0.147\linewidth]{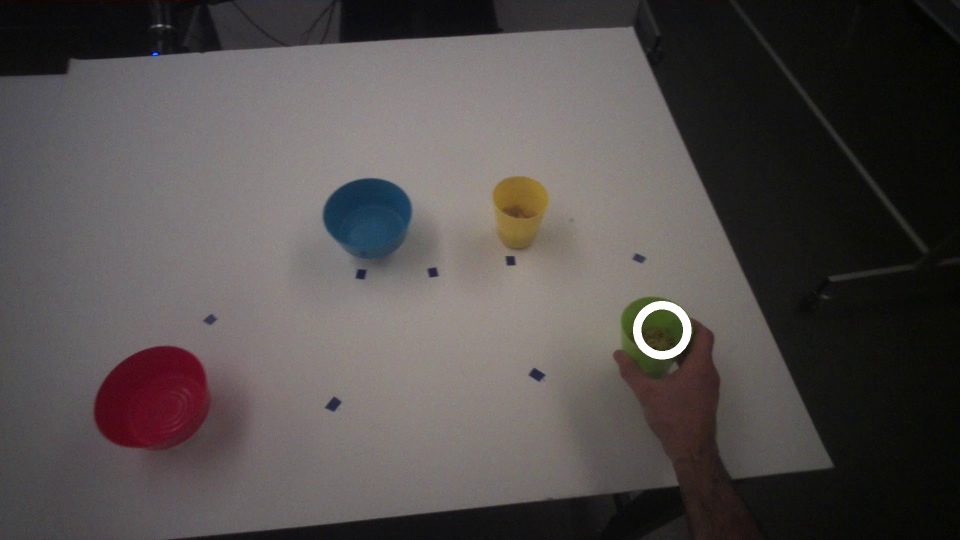}
}
\subfigure[Pour %\newline Target:Red Bowl
]{
\includegraphics[width=0.147\linewidth]{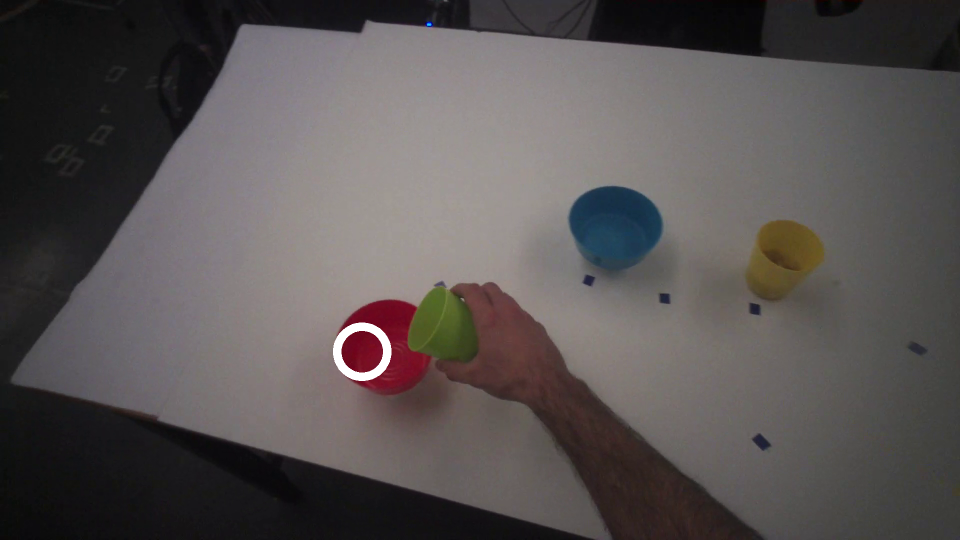}
}
\subfigure[Return  %\newline  Target:White Table
]{
\includegraphics[width=0.147\linewidth]{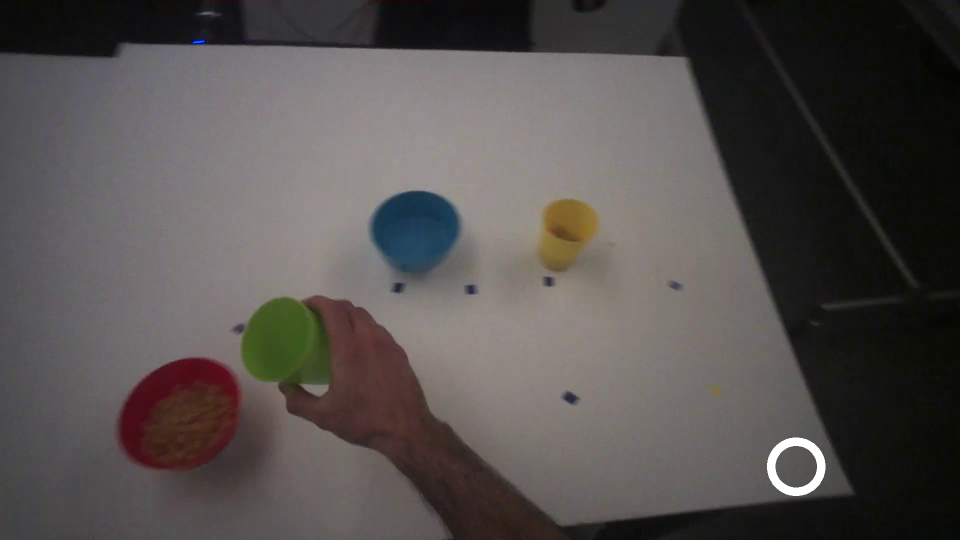}
}
\subfigure[Release %\newline Target:Green Cup
]{
\includegraphics[width=0.147\linewidth]{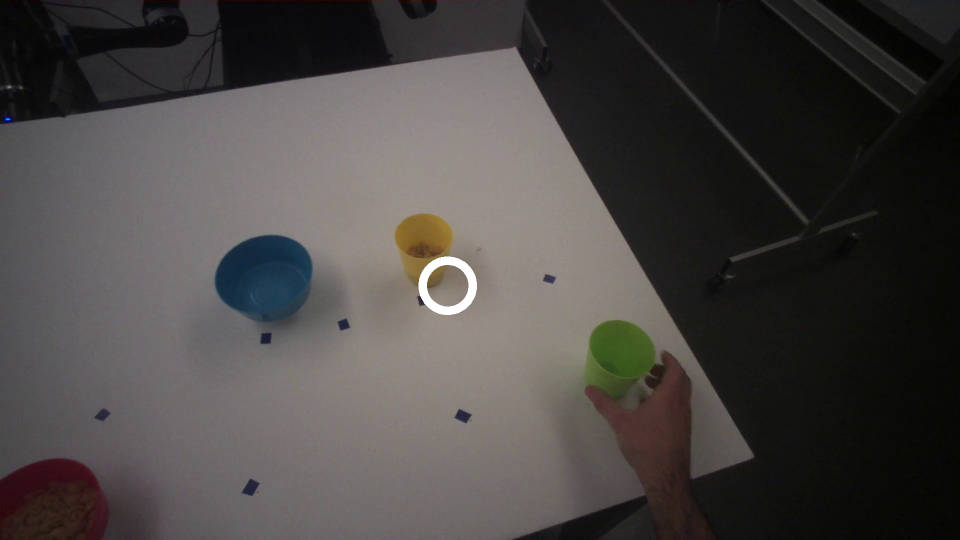}
}
\caption{Semantic keyframes at the start of each pouring subtask with corresponding gaze fixation points (white circle). The reference frames for these subtasks are (i) green cup  for reach, grasp, release; (ii) yellow cup for reach, grasp, release; (iii) red bowl for transport and pour; (iv) blue bowl for transport and pour; (v) white table  for return.}
% \vspace{-5mm}
\label{fig:keyframes}
\end{figure*}

Users wore the Tobii Pro Glasses 2 eye tracker and provided demonstrations to our robot which has a 7 degree of freedom (DOF) Kinova arm, and a Kinect sensor mounted on its head. The tracker technology ensures automatic compensation for slipping and makes it possible to track eye gaze reliably in dynamic environments. It has a simple calibration process in which users stare for a few seconds at the center of a calibration card with concentric circles, provided by the manufacturer. The first-person view and corresponding eye tracking data from the glasses (Fig. \ref{fig:keyframes}) are recorded at 50 Hz. The data is saved onto a pocket-sized recording unit that allows the participant to move around unrestricted. The eye tracker was individually calibrated once at the beginning of the study for each user, and subsequently calibrated in-between demonstrations if there was data loss over the network or the user wanted to take a break. For two users the tracker did not calibrate well after multiple trials, hence only a part of the study could be conducted with them. Due to noisy observations or loss of data transmission, we eliminated data from such users, which left us with 8 expert and 8 novice users for the pouring task and 9 novice and 7 expert users for the placement task. This amounted to a total of $\sim$27 minutes and $\sim$124 minutes of video and KT demonstration data, respectively.

KT demonstrations required users to physically move the robot's arm while video demonstrations were given standing in front of the robot, in the robot camera's view (Fig. \ref{fig:tasks}(c)). For KT demonstrations, keyframes are provided by the users themselves. However, they do not attach a semantic meaning to them. We annotate each keyframe with one of the subtask labels shown in Fig. \ref{fig:keyframes}. These keyframes are then synchronized with the gaze data time stamps to recover at what point in the first person video the keyframes lie. Users provide different levels of granularity in their keyframe segmentations. For example, some users break down the pouring action into multiple keyframes in which the gripper is being rotated at different angles until all the pasta falls out, and some only rotate the wrist joint once and mark the end of the pouring action as a keyframe. Video demonstrations are relatively shorter in duration, as the user is able to complete the task within seconds. Therefore, we manually annotate videos with a fixed number of semantically meaningful keyframes for the pouring task. Examples of keyframes for the pouring task in a video demonstration type are shown in Fig. \ref{fig:keyframes}. The placement task is relatively simple, as it only requires logging a single action of placing the ladle on the table.

\subsection{Gaze Filtering}
The eye tracker is equipped with two cameras for each eye to track gaze and one scene camera to record what the user sees. We collected the following data at 50 Hz: (1) raw world camera images in the user's egocentric view, (2) pixel location of the human's gaze in the egocentric image, (3) gaze time stamps synchronized with keyframe time stamps along the KT demonstration. Raw gaze locations are processed to extract spatio-temporal features of gaze such as fixations and saccades. %, and smooth pursuits %\scott{Not just *can* -- you actually did, right? Say so.} 
\cite{kasneci2017aggregating, nystrom2010adaptive}. Eye gaze movements can be characterized as: (a) Fixations, (b) Saccades, (c) Smooth pursuits, and (d) Vestibulo-ocular movements. Visual fixations maintain the focus of gaze on a single location. Fixation duration varies based on the task, but one fixation is typically 100 - 500 ms, although it can be as short as 30 ms \cite{holmqvist2011eye}. Saccades are rapid, ballistic,  voluntary eye movements (usually between 20 - 200 ms) that abruptly change the point of fixation. Smooth pursuit movements are slower tracking movements of the eyes that keep a moving stimulus on the fovea. Such movements are voluntary in that the observer can choose to track a moving stimulus, but only highly trained people can make smooth pursuit movements without a target to follow. Smooth pursuit movements are minimally present  in  our  trials  and  are  preserved  after  filtering  for  saccades. Vestibulo-ocular  movements stabilize the eyes relative to the external world to compensate for head movements.  These reflex responses prevent visual images from slipping on the surface of the retina as head position changes. The accelerometer and gyroscope sensors of the eye tracker glasses differentiate between head and eye movements which eliminates the impact of head movements on eye tracking data.

Salvucci et al. \cite{salvucci2000identifying} proposed a novel taxonomy of fixation identification algorithms and evaluated existing algorithms in the context of this taxonomy. They identify two characteristics---spatial and temporal--- to classify different algorithms for fixation identification. For spatial characteristics, three criteria distinguish primary types of algorithms: velocity-based, dispersion-based, and area-based. For temporal characteristics, they include two criteria: whether the algorithm uses duration information, and whether the algorithm is locally adaptive. The use of duration information is guided by the fact that fixations are rarely less than 100 ms and often in the range of 100-500 ms. In our work, we use velocity-based and area-based criteria under spatial characteristics and duration based criteria under temporal characteristics to filter out fixations from saccades. We first filter out eye movements with very high speeds (a large distance traversed over a very short period of time is likely a saccade). Then we compute object color histograms in a circular area (100-pixel radius) around the 2D eye gaze location obtained from the eye tracker. The object is identified as the focus of attention for that instant if the color of the object is present in a majority of the pixels around the gaze point detected by the eye tracker, since all objects in our tasks have significantly different colors. If gaze remains on one such object for more than 100 ms, we consider it a fixation.

\begin{figure}
\centering
%\vspace{2mm}
\subfigure[Pouring Task]{
\includegraphics[width=0.23\textwidth]{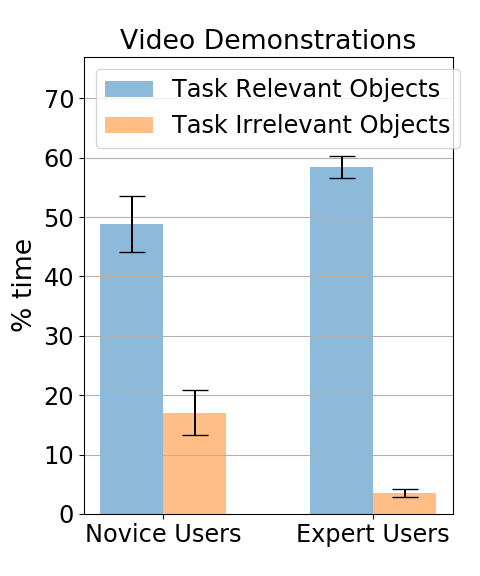}
\includegraphics[width=0.23\textwidth]{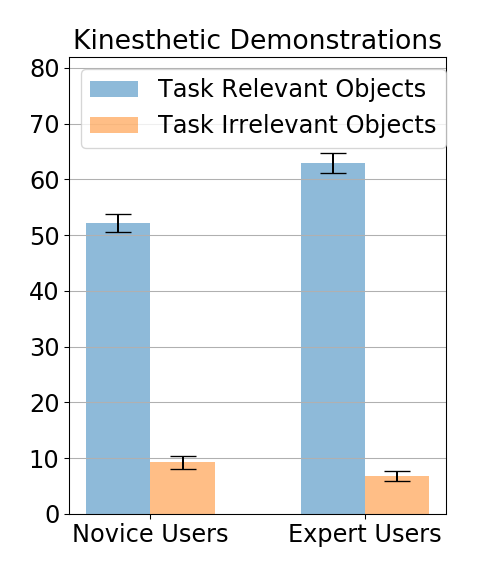}
}
\subfigure[Placement Task]{
\includegraphics[width=0.23\textwidth]{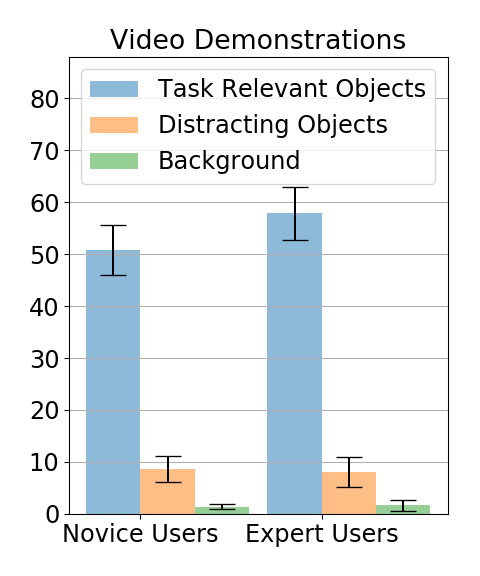}
\includegraphics[width=0.23\textwidth]{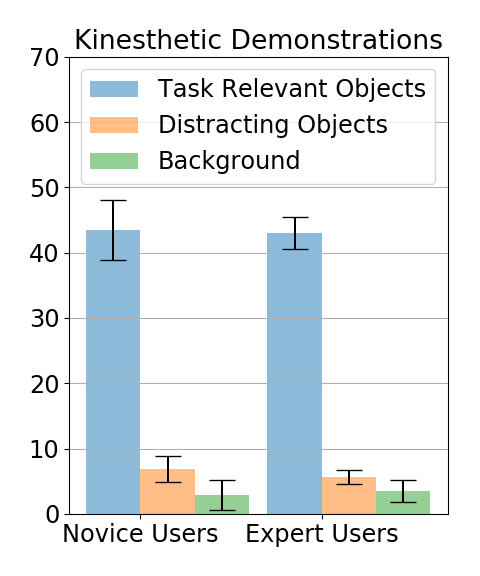}
}
\caption{Avg. \% time spent by novice \& expert users fixating on task-relevant v/s irrelevant objects.}
% \vspace{-5mm}
\label{fig:results}
\end{figure}

%===============================================================================

\section{Experiments and Results}
\subsection{Statistical Analysis of Gaze Patterns for LfD}
\textbf{Users rarely fixate on task-irrelevant objects: }
Consistent with prior work \cite{mennie2007look} we find that under both tasks and both forms of demonstrations, users fixate more on objects which are relevant to the task. Specifically for the pouring task, the two-way mixed design ANOVA test produces $F (1,46) = 100.94, p<0.01$ (video demonstrations) and $F(1,40) = 762.80, p<0.01$ (KT demonstrations) showing task relevance of objects of gaze fixation. A large $F$ ratio implies that the variation among group means is more than expected to be seen by chance. The $p$ value is computed from the $F$ ratio, which tests the null hypothesis that data from all groups are drawn from populations with identical means. 
The main effect is significant for both demonstration types ($p<0.01$; Fig. \ref{fig:results}), i.e. gaze fixations on task-relevant and task-irrelevant objects come from different distributions. In KT demonstrations, there is a significant difference in fixation duration between user type ($F(1,40)=20.39, p<0.01$) and significant interaction effect between task-relevance and user type ($F(1,40)=13.62, p<0.01$). For the placement task as well, significant differences are observed between fixation duration on task relevant and task-irrelevant objects ($p<0.01$) for each demonstration type. This provides strong evidence that using gaze during demonstrations can help to identify the relative importance of different parts of the workspace.

\begin{figure*}
\centering
%\vspace{2mm}
\subfigure[Expert Users]{
\includegraphics[width=0.44\textwidth]{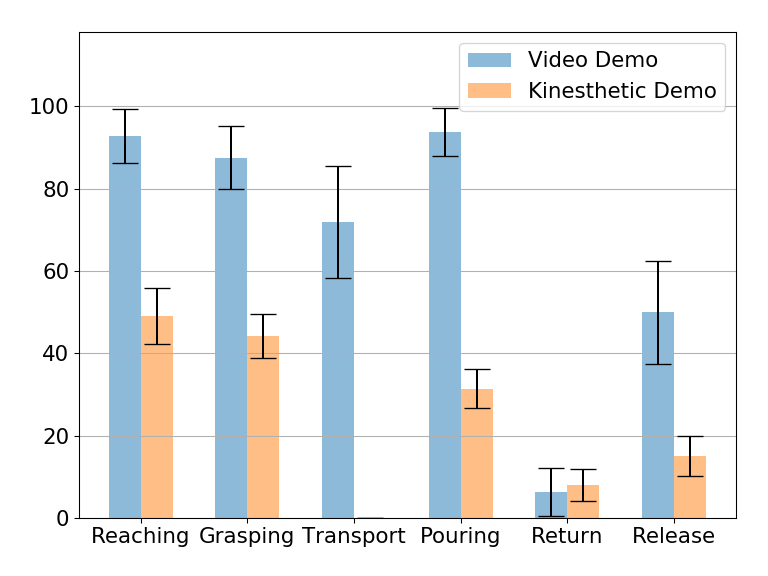}
}
\subfigure[Novice Users]{
\includegraphics[width=0.44\textwidth]{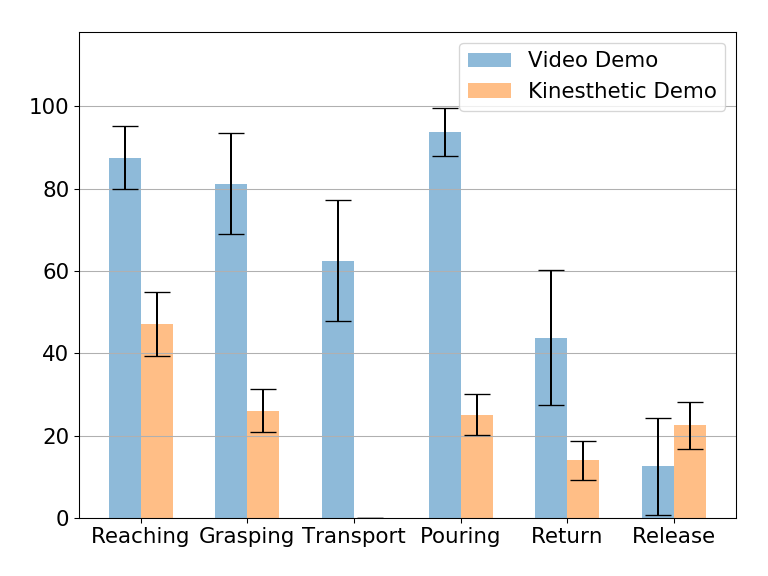}
 }
\caption{Reference frame detection accuracy for each action of the pouring task under two demonstration types (video demonstrations and keyframe-based kinesthetic demonstrations).}
% \vspace{-5mm}
\label{fig:target}
\end{figure*}

\textbf{User fixations can predict the target object of keyframes for video demonstrations:}
\label{sec:clean}
Video demonstrations contain cleaner gaze fixation patterns than KT demonstrations, where object fixations are interspersed with glances at the gripper: the total number of consecutive object-fixation changes across all KT demonstrations for the pouring task are $12\times$ higher than that for video demonstrations. We computed fixation patterns between distinct keyframes for the first trial of video (manually coded) and KT demonstrations (user provided) for each user. In video demonstrations, fixations between keyframes (Fig. \ref{fig:keyframes}) signifying the semantic action of reaching, grasping, transport and pouring line up with their target reference frames at least 75\% of the time for expert users and at least 70\% of the time for novice users (Fig. \ref{fig:target}). 

\begin{figure*}
\centering
%\vspace{2mm}
\subfigure[All Users (VD)]{
\includegraphics[width=0.23\textwidth]{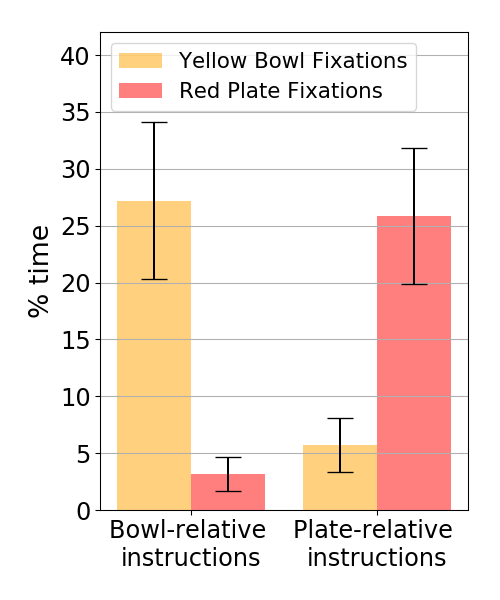}
}
\subfigure[Expert Users (KD)]{
\includegraphics[width=0.23\textwidth]{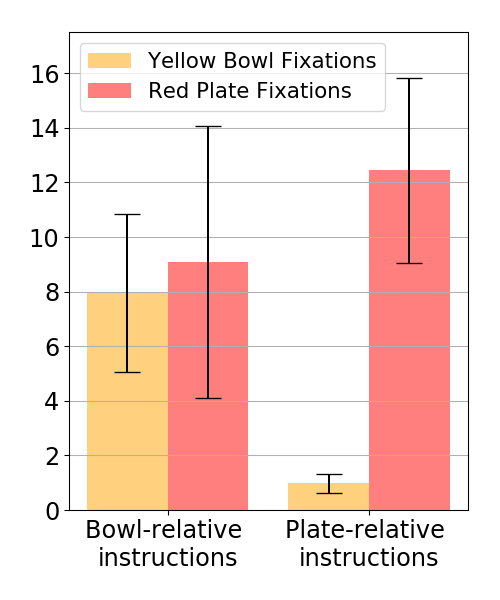}
}
\subfigure[Novice Users (KD)]{
\includegraphics[width=0.23\textwidth]{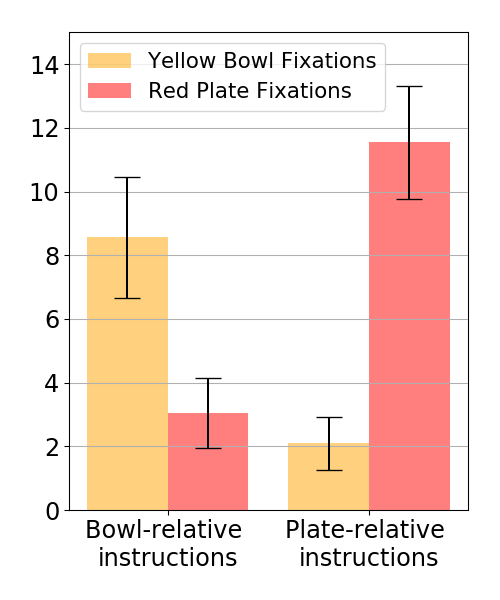}
}
\subfigure[Gaze on Robot Gripper]{
\includegraphics[width=0.23\textwidth]{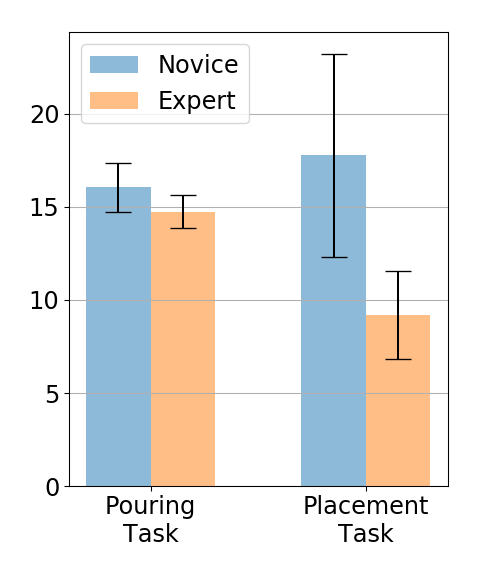}
}
\caption{\% of time spent fixating on the red plate and yellow bowl during placement demonstrations for (a) all users during video demonstrations (VD), (b) expert users and (c) novice users during KT demonstrations (KD). (d) Proportion of time by user expertise spent fixating on the gripper relative to task objects in the pouring and placement tasks during KT demonstrations.}
\label{fig:reward-placement}
\end{figure*}

\textbf{Novice robot users attend more to the robot's gripper: }
%Compared to Experts
For KT demonstrations of the pouring task, we find that novice users on average spend more time fixating on the gripper than expert robot users (Fig. \ref{fig:reward-placement}(d)), likely because novice users often struggle to manipulate the robot's arm, and thus focus more on moving the gripper. Even though the results are not statistically significant across the entire pouring task ($p=0.813$) or when observing a single action for the placement task ($p=0.178$), an overall average difference exists between user types. Both expert and novice users independently still spend more time overall on the task-relevant objects compared to the gripper. 

\textbf{Gaze can identify intent for ambiguous actions: }
In the placement task, the ladle is placed at roughly the same location on the table for 2 instructions (left of the red plate or right of the yellow bowl) given to the user (Fig. \ref{fig:tasks}(b)). A different spatial reference frame for placing the ladle should change the user's internal objective function and we expect this to be reflected in the amount of time spent fixating on the object representing the reference frame. For the instruction relative to the red plate, we find that all users on average fixate more on the plate in comparison to the bowl in both video and KT demonstrations (Fig. \ref{fig:reward-placement}). They similarly fixate relatively more on the yellow bowl for the instruction relative to the bowl. Our results for video demonstrations and for novice users of KT demonstrations are statistically significant ($p<0.01$). This finding aligns with past research on understanding gaze for natural manipulation behavior, showing strong promise to be utilized as an additional signal for inferring reward functions from demonstrations.

\textbf{Gaze patterns differ between step and non-step keyframes: }
We refer to keyframes of KT demonstrations which mark the boundaries of semantically different actions (such as Fig. \ref{fig:keyframes}) as step keyframes. We hypothesize gaze fixations before and after such keyframes will more likely constitute different objects of attention compared to non-step keyframes. The object of attention on which a user spends the maximum time fixating 3 seconds before and 3 seconds after every keyframe is computed. For novice users, the target object of attention is different before and after 19.44\% of non-step keyframes, and 24.51\% of step keyframes. For expert users, the target object of attention is different before and after 15.79\% of non-step keyframes, and 27.85\% of step keyframes. This implies an average of 6\% and 12\% more of step keyframes constitute a change in the object of attention compared to non-step keyframes for novice and expert users respectively. Even though gaze alone might not be sufficient in distinguishing between step and non-step keyframes, gaze can be a useful feature in addition to other features for this classification task. We propose the use of gaze for keyframe classification as an open problem.

%===============================================================================

\subsection{Utilizing Human Gaze for Learning}
\subsubsection{Subtask Prediction}

Many LfD methods focus on the case in which the robot learns a monolithic policy from a demonstration of a simple task with a well-defined beginning and end \citep{argall2009survey}. However, this approach often fails for complex tasks that are  difficult  to  model  with  a  single policy. 
Several household tasks require multiple steps comprising of different actions involving different goals, objects and features. It is, therefore, important to segment a complex task into simpler subtasks, and then learn subsequent policies for each subtask. It has been shown that learning a separate policy for each step of a task can lead to improved generalization \citep{niekum2012learning, konidaris2012robot}.

\begin{figure*}
\centering
\includegraphics[width=1.0\textwidth]{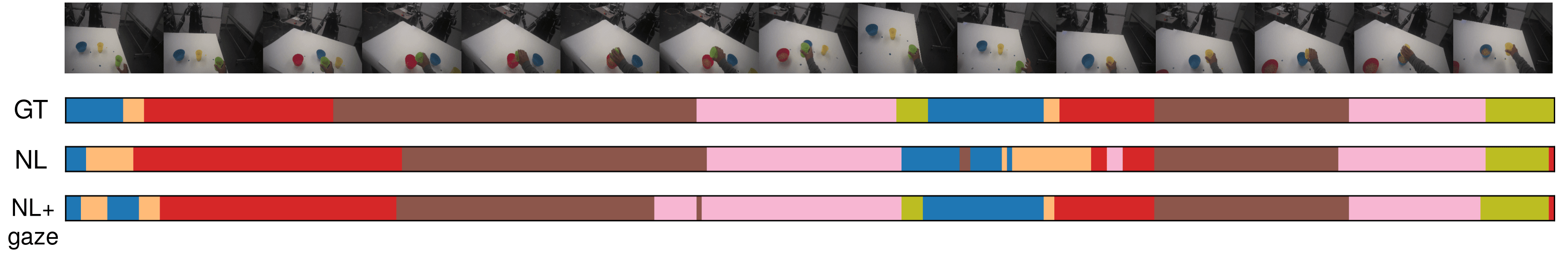}
\caption{Visualization of sub-task prediction for a video demo of the pouring task. Each color represents a different subtask label. The rows below the demonstration images show ground truth labels (GT), labels from a non-local neural network not using gaze (NL), labels from a non-local neural network using gaze (NL + gaze).}
% \vspace{-5mm}
\label{fig:seg}
\end{figure*}

Gaze fixation patterns accumulated and analyzed over subtasks reveal that gaze can predict their target reference frames well, especially for video demonstrations (Section \ref{sec:clean}). Motivated by this finding, we show in a proof-of-concept experiment that gaze can improve automatic subtask classification for multi-step demonstrations, as an intermediate step to multi-step policy learning.   
We use two different model architectures for subtask classification: (i) Non-Local (NL) neural network \citep{wang2018non} and (ii) Compact Generalized Non-Local (CGNL) neural network \citep{CGNLNetwork2018}, which have been shown to work well for activity recognition. We use the ResNet-50 architecture with one NL or one CGNL block, weights initialized from a pre-trained ImageNet model. To incorporate gaze information, we use the normalized gaze coordinates as input before the last fully connected layer for both these networks. Egocentric videos from the eye tracker are sub-sampled to generate $\sim$3K images for video demonstrations and $\sim$12K images for KT demonstrations. In a 10-fold cross validation experiment, action labels are predicted per frame by each model.  
Incorporating gaze improves accuracy of subtask prediction with both models for both demonstration types (Table \ref{tab:action-classification}). An example of the action labels predicted is shown in Fig. \ref{fig:seg}. Even though action classification at the snippet-level does not necessarily yield clean, contiguous activity segment labels, Goo et al. \citep{gooone} showed that even with moderate action label noise, reward inference and subsequent policy learning can improve versus policy learning on entire videos without any segmentation.

\begin{table*}[!htb]
    \centering
    \begin{tabular}{|c|c|c||c|c|}
    \hline
    & \multicolumn{2}{c||}{Video Demos} & \multicolumn{2}{c|}{KT Demos}\\
    \hline
         &  NL & CGNL  &  NL & CGNL \\
        \hline
        Without Gaze & 86.95 & 82.42 & 61.95 & 63.00\\
        \hline
        With Gaze & \textbf{87.63} & \textbf{88.88} & \textbf{62.71} & \textbf{64.11}\\
        \hline
    \end{tabular}
    \caption{10-fold cross validation accuracy of subtask prediction. 
    }
    % \vspace{-5mm}
    \label{tab:action-classification}
\end{table*}

\subsubsection{Reward Learning}
\label{sec:placementanalysis}
We hypothesize that differences in the amount of time spent looking at an object of interest can arise from the intent or internal reward of the demonstrator. The role of internal reward in guiding eye and body movements has been observed in neurophysiological studies \citep{hayhoe2005eye}. 
Specifically, neural recordings have shown that vast areas of the brain's gaze computation system exhibit sensitivity to reward signals \citep{hayhoe2005eye}. 
To investigate this hypothesis, we examine the possible role of gaze in an inverse reinforcement learning (IRL) setting. IRL offers an intuitive means to specify robot goals by providing demonstrations from which the robot can recover the reward function to optimize. One method for IRL is Bayesian inverse reinforcement learning (BIRL) \citep{ramachandran2007bayesian}, which models the posterior distribution, $P(R|D) \propto P(D|R)P(R)$, over reward functions $R$, given demonstrations $D$. BIRL estimates the probability of state-action pairs of a demonstration set, given a reward, to infer this distribution. It assumes the demonstrator follows a softmax policy, resulting in the following likelihood function:
 
\begin{equation}
    P(D|R) = \prod_{(s,a) \in D} \frac{e^{cQ^*_R(s,a)}}{\sum_{b \in A}e^{cQ^*_R(s,b)}}
\end{equation}

\noindent where $c$ is a parameter representing the degree of confidence we have in the demonstrator's ability to choose the optimal actions \cite{ramachandran2007bayesian}, and $Q^*_R$ denotes the Q-function of the optimal policy under reward $R$. Markov Chain Monte Carlo (MCMC) sampling is used to obtain samples from the posterior, from which an estimate of the maximum a posteriori (MAP) reward function $R_{MAP}$ or the mean reward function $\overline{R}$ can be extracted.

Additional information, such as the gaze of the demonstrator, is typically ignored in IRL algorithms. We recover the reward function for different instructions of the placement task (placement with respect to red plate or yellow bowl) using gaze with BIRL. By incorporating gaze information $G$ as a prior into this framework, we formulate the posterior as follows: %$P(R|D,G) = P(D|R)P(R|G)$, 

\begin{equation}
    P(R|D,G) \propto P(D|R)P(R|G)
\end{equation}

where we model $P(R|G) = -\sum_{i,j}I_{ij}\frac{f_i}{f_j}$ and $I_{ij}$ is an indicator function which is $1$ when $w_i<w_j$ and $f_i>f_j$. $f_i$ is the time spent fixating at object $i$ and $w_i$ is the sum of the 5 RBF kernel weights: 4 RBFs are placed around (top-right, top-left, bottom-right, bottom-left) and 1 is placed at the center of the object $i$. The RBFs are used to capture spatial information relative to objects \cite{brown2019risk}, and the ratio of fixation times captures relative attention given to objects. The indicator function is 1 if the ranking of RBF weight magnitudes for a pair of objects does not match the ranking of the magnitude of fixation times on the respective objects. 
Given k items of interest on the table, we assume the reward for placement location x is given by:
\begin{equation}
R(x) = \sum_{i=1}^{k} \sum_{j=1}^5 w_{ij} \cdot rbf(x,c_{ij},\sigma_i^2)
\end{equation} 

with \begin{equation} rbf(x,c,\sigma^2)=exp(-{\left|\left|x-c\right|\right|}^2/\sigma^2). \end{equation} 

\begin{comment}
\begin{equation}
    \nabla_xR(x) = \sum_{i=1}^k\sum_{j=1}^5w_{ij} \cdot rbf(x,c_{ij},\sigma_i^2) \bigg( \frac{-2x+2c_{ij}}{\sigma_i^2}\bigg)
\end{equation}
\end{comment}

This formulation downweights reward functions in which the relative time spent fixating near two objects does not match the relative weights assigned to their RBF kernels (i.e. we expect features to have larger magnitude weights and influence the reward function more when they are defined relative to objects that were looked at more frequently). 

We hypothesize that incorporating gaze in BIRL will help to identify the important object-relations for the task, thereby imposing preferences over reward functions that might otherwise appear equally good when looking at demonstrations without gaze information. We find an improvement in policy learning (Table \ref{tab:birl-policy}, \ref{tab:birl-place}) after performing reinforcement learning on the inferred reward function. Gaze fixation times on the yellow bowl and red plate across 5 ambiguous video and KT demonstrations are used to determine how well incorporating the fixation time performs relative to ignoring the gaze information (standard BIRL algorithm). Given the instruction for placement, we formulate a ground truth reward in which the ladle should be placed as instructed (e.g.: for the instruction to place the ladle on the right of the bowl, weights of RBFs on the top-right and bottom-right of the bowl are set to 0.5 each and all remaining RBF weights are set to 0). With the demonstrated placement location and gaze fixation time, the underlying reward function is recovered. The placement policy is computed by picking the best position via gradient ascent with random restarts. Generalization is measured under 100 different configurations of the bowl and plate in simulation. 

To evaluate our experiment, we use two metrics: the policy loss and placement loss. The policy loss of executing a policy $\pi$  under the reward $R$ is
given by the Expected Value Difference:
\begin{equation}
    EVD(\pi , R) = V_R^{\pi^*} - V_R^{\pi}
\end{equation}

% \begin{equation}
%     V^{\pi_1}_{R^*} - V^{\pi_2}_{R_{MAP}}
% \end{equation}

% \begin{equation}
%     |x^* - x_{R_{MAP}}| 
% \end{equation}

$\pi = \pi_{MAP}$ is used as the robot's best guess of the optimal policy under the demonstrator's reward, where $\pi_{MAP}$ is the optimal policy corresponding to $R_{MAP}$, the maximum a posteriori reward given the demonstrations so
far. $\pi^*$ is the optimal policy corresponding to the ground truth reward. The placement loss is computed using the difference between the ground truth placement location and the placement location estimated by $\pi_{MAP}$. We find that both policy loss and placement loss are lower when gaze is incorporated into the learning framework. Even with a single ambiguous demonstration, gaze improves performance.  Since video demonstrations contain richer gaze signals (Sec. \ref{sec:clean}), there is an overall greater improvement in both metrics when incorporating gaze from video demonstrations. 
We envision that the use of gaze information in other learning algorithms would also result in better generalization performance, which we pose as an open problem for future work.

\begin{table*}[!htb]
    \centering
    \begin{tabular}{|c|c|c|c|c|c|c|c|c|}
    \hline
         &  \multicolumn{2}{c|}{5 KT Demos} & \multicolumn{2}{c|}{5 Video Demos} &  \multicolumn{2}{c|}{1 KT Demo} & \multicolumn{2}{c|}{1 Video Demo}\\
        \hline
        Instruction relative to  & Bowl & Plate & Bowl & Plate & Bowl & Plate & Bowl & Plate\\
        \hline
        Without Gaze & 0.619 & 0.081 & 0.678 & 0.036 & 0.073 & 0.666 & 0.486 & 0.184\\
        \hline
        With Gaze & \textbf{0.329} & \textbf{0.032} & \textbf{0.046} & \textbf{0.021} & \textbf{0.043} & \textbf{0.225} & \textbf{0.098} & \textbf{0.120} \\
        \hline
        \hline 
        Avg improvement w/ Gaze & \multicolumn{2}{c|}{53.7\%} & \multicolumn{2}{c|}{\textbf{67.4\%}} &  \multicolumn{2}{c|}{53.6\%} & \multicolumn{2}{c|}{\textbf{57.3\%}}\\
        \hline
    \end{tabular}
    \caption{Average policy loss w/ and w/o gaze information in BIRL for the placement task.}
    % \vspace{-5mm}
    \label{tab:birl-policy}
\end{table*}

\begin{table*}[!htb]
    \centering
    \begin{tabular}{|c|c|c|c|c|c|c|c|c|}
    \hline
         &  \multicolumn{2}{c|}{5 KT Demos} & \multicolumn{2}{c|}{5 Video Demos} &  \multicolumn{2}{c|}{1 KT Demo} & \multicolumn{2}{c|}{1 Video Demo}\\
        \hline
        Instruction relative to  & Bowl & Plate & Bowl & Plate & Bowl & Plate & Bowl & Plate\\
        \hline
        Without Gaze & 0.494 & 0.102 & 0.536 & 0.064 & 0.087 & 0.492 & 0.383 & 0.160\\
        \hline
        With Gaze & \textbf{0.291} & \textbf{0.063} & \textbf{0.068} & \textbf{0.045} & \textbf{0.066} & \textbf{0.191} & \textbf{0.102} & \textbf{0.122} \\
        \hline
        \hline 
        Avg improvement w/ Gaze & \multicolumn{2}{c|}{39.7\%} & \multicolumn{2}{c|}{\textbf{58.5\%}} &  \multicolumn{2}{c|}{42.7\%} & \multicolumn{2}{c|}{\textbf{48.6\%}}\\
        \hline
    \end{tabular}
    \caption{Average placement loss w/ and w/o gaze information in BIRL for the placement task.}
    % \vspace{-3mm}
    \label{tab:birl-place}
\end{table*}

%=========================================================================

\section{Discussion and Conclusions}
In this work, we showed that human gaze behavior during teaching is informative in a variety of ways. We find that gaze behaviors exhibited during video demonstrations and KT demonstrations are similar in that users mostly fixate on objects being manipulated or objects with respect to which manipulation occurs. These demonstration modalities differ in terms of gaze fixations of video demonstrations lining up more accurately with target objects for a semantic action, and being more informative to improve reward inference with Bayesian IRL for a simple placement task. Consistent with previous findings is the notion that gaze of a user reflects their internal reward function. Particularly during ambiguous demonstrations, when it is unclear from a single demonstration what feature in the workspace the user is trying to optimize, gaze can reveal intentions which are not directly observable from the action alone. 

We also discover eye gaze patterns specific to demonstrations for robots. Specifically, human gaze fixations during demonstrations differ for step and non-step keyframe segments; and between users with different robot-expertise. Also, gaze fixations can identify target objects of subtasks part of a multi-step video demonstration. Motivated by this finding, we show utilizing gaze leads to an improvement in subtask classification from egocentric videos of both demonstration types. Most importantly, our results show an existence proof on the informativeness of gaze data and related open problems for the research community. We envision that gaze information will increasingly be used to improve applications including automatic task segmentation, policy learning, video and kinesthetic demonstration alignment, keyframe classification, and reference frame detection.

Hayhoe et al. \cite{hayhoe2005eye} note that there is a pervasive role of the task itself in guiding where and when to fixate. The gaze patterns we observed are tied to the goal-oriented nature of the tasks performed and might vary with added complexity of the task. Considering more complex tasks which require search and planning, along with tasks where trajectories matter %and not the end goal 
will be interesting avenues to better understand gaze behavior for teaching. Moreover, in this work we make use of eye trackers to observe the gaze of a human accurately during a  demonstration. We recognize that additional eye tracking equipment may not be available to end users at all times and comes with a cost for improving the performance of robot learning algorithms. We note that there are several vision-based gaze tracking algorithms \cite{chong2018connecting, recasens2015they}, usable in real time by robots to infer human gaze fixations. In future work, such solutions can be used to detect gaze fixations and incorporate them in learning algorithms.

%===============================================================================

\acknowledgments{This work has taken place in the Personal Autonomous Robotics Lab (PeARL) and the Socially Intelligent Machines (SIM) Lab at The University of Texas at Austin. PeARL research is supported in part by the NSF (IIS-1724157, IIS-1638107, IIS-1749204). SIM research is supported in part by NSF (IIS 1724157, IIS 1638107) and ONR (N000141612835, N000141612785). We thank Dr. Garrett Warnell (Army Research Laboratory and University of Texas at Austin) for access to the Tobii Pro Glasses 2 eye tracker and Ruohan Zhang (University of Texas at Austin) for advice about it's use.}

%===============================================================================

\bibliography{main}  % .bib

%\clearpage
\end{document}